\documentclass[10pt, a4paper]{article}
\usepackage{lrec}
\usepackage{multibib}
\newcites{languageresource}{Language Resources}
\usepackage{tabularx}
\usepackage{soul}
\usepackage{times}
\usepackage{url}
\usepackage{latexsym}
\usepackage{graphicx}
\usepackage{amsmath}
\usepackage{multirow}
\usepackage{graphicx}

\usepackage{epstopdf}
\usepackage[latin1]{inputenc}

\usepackage{hyperref}
\usepackage{xstring}

\title{Indian Language Wordnets and their Linkages with Princeton WordNet}

\name{Diptesh Kanojia$^{1,2,3}$, Kevin Patel$^{1}$, Pushpak Bhattacharyya$^{1}$}

 \address{$^{1}$IIT Bombay, $^{2}$Monash University, \\
         $^{3}$IITB-Monash Research Academy,\\
          \{diptesh, kevin.patel, pb\}@cse.iitb.ac.in\\}
\abstract{
Wordnets are rich lexico-semantic resources. Linked wordnets are extensions of wordnets, which link similar concepts in wordnets of different languages. Such resources are extremely useful in many Natural Language Processing (NLP) applications, primarily those based on knowledge-based approaches. In such approaches, these resources are considered as gold standard/oracle. Thus, it is crucial that these resources hold correct information. Thereby, they are created by human experts. However, human experts in multiple languages are hard to come by. Thus, the community would benefit from sharing of such manually created resources. In this paper, we release mappings of 18 Indian language wordnets linked with Princeton WordNet. We believe that availability of such resources will have a direct impact on the progress in NLP for these languages.}

\begin{document}

\maketitleabstract

\section{Introduction}

Wordnets \cite{fellbaum1998wordnet} have been useful in different Natural Language Processing applications such as Word Sense Disambiguation \cite{tufics2004fine,sinha2006approach}, Machine Translation \cite{knight1994building} \textit{etc.} 

Linked Wordnets are extensions of wordnets. In addition to language-specific information captured in constituent wordnets, linked wordnets have a notion of an interlingual index, which connects similar concepts in different languages. Such linked wordnets have found their application in machine translation \cite{hovy1998combining}, cross-lingual information retrieval \cite{gonzalo1998indexing}, \textit{etc.}

Given the extensive application of wordnets in different NLP applications, creation and maintenance of wordnets involve expert involvement. Such involvement is costly both in terms of time and resources. This is further amplified in case of linked wordnets, where experts need to have knowledge of multiple languages. 

India is a vast country with massive language diversity. According to a census in 2001, there are 122 major languages \footnote{http://en.wikipedia.org/wiki/Languages\_of\_India}, out of which, 29 have more than a million native speakers. The IndoWordNet project contains wordnets of 18 of these languages. These wordnets were created using expansion approach with Hindi Wordnet as the pivot. 

This paper makes the following contributions:

\begin{itemize}
	\item We release the latest version of 18 wordnets under the IndoWordNet project as a single bundle.
    \item Using mappings between Princeton WordNet and Hindi wordnet, we create and release mappings between Princeton WordNet and these 18 languages wordnet.
\end{itemize}

The rest of the paper is organized as follows: Section \ref{bg} covers some background and related work needed for further discussions. Section \ref{resources} describes the released resources. Section \ref{disc} discusses different issues encountered in the creation of these datasets, followed by the conclusion and future work.

\section{Background and Related Work}
\label{bg}
Princeton WordNet or the English WordNet was the first wordnet and inspired the development of many other wordnets. EuroWordNet \cite{1997Vossen} is a linked wordnet comprising of wordnets for European languages, \textit{viz}, Dutch, Italian, Spanish, German, French, Czech and Estonian. Each of these wordnets is structured in the same way as the Princeton WordNet for English \cite{1990Miller} - synsets (sets of synonymous words) and semantic relations between them. Each wordnet separately captures a language-specific information. In addition, the wordnets are linked to an Inter-Lingual-Index, which uses Princeton WordNet as a base. This index enables one to go from concepts in one language to similar concepts in any other language. Such features make this resource helpful in cross-lingual NLP applications.

IndoWordNet \cite{2010Bhattacharyya} is a linked wordnet comprising of wordnets for major Indian languages, \textit{viz}, Assamese, Bengali, Bodo, Gujarati, Hindi, Kannada, Kashmiri, Konkani, Malayalam, Manipuri, Marathi, Nepali, Oriya, Punjabi, Sanskrit, Tamil, Telugu, and Urdu. These wordnets have been created using the expansion approach with Hindi WordNet as a pivot, which is partially linked to English WordNet. We exploit these links to create mappings from English WordNet to wordnets of other languages.

\begin{table*}[]
\centering
\resizebox{0.70\textwidth}{!}{%
\begin{tabular}{c|r|r|r|r|r|}
\cline{2-6}
 & \multicolumn{1}{c|}{Noun} & \multicolumn{1}{c|}{Verb} & \multicolumn{1}{c|}{Adjectives} & \multicolumn{1}{c|}{Adverbs} & \multicolumn{1}{c|}{Total} \\ \hline
\multicolumn{1}{|c|}{Assamese} & 9065 & 1676 & 3805 & 412 & 14958 \\ \hline
\multicolumn{1}{|c|}{Bengali} & 27281 & 2804 & 5815 & 445 & 36346 \\ \hline
\multicolumn{1}{|c|}{Bodo} & 8788 & 2296 & 4287 & 414 & 15785 \\ \hline
\multicolumn{1}{|c|}{Gujarati} & 26503 & 2805 & 5828 & 445 & 35599 \\ \hline
\multicolumn{1}{|c|}{Hindi} & 29807 & 3687 & 6336 & 541 & 40371 \\ \hline
\multicolumn{1}{|c|}{Kannada} & 12765 & 3119 & 5988 & 170 & 22042 \\ \hline
\multicolumn{1}{|c|}{Kashmiri} & 21041 & 2660 & 5365 & 400 & 29469 \\ \hline
\multicolumn{1}{|c|}{Konkani} & 23144 & 3000 & 5744 & 482 & 32370 \\ \hline
\multicolumn{1}{|c|}{Malayalam} & 20071 & 3311 & 6257 & 501 & 30140 \\ \hline
\multicolumn{1}{|c|}{Manipuri} & 10156 & 2021 & 3806 & 332 & 16351 \\ \hline
\multicolumn{1}{|c|}{Marathi} & 23271 & 3146 & 5269 & 539 & 32226 \\ \hline
\multicolumn{1}{|c|}{Nepali} & 6748 & 1477 & 3227 & 261 & 11713 \\ \hline
\multicolumn{1}{|c|}{Odiya} & 27216 & 2418 & 5273 & 377 & 35284 \\ \hline
\multicolumn{1}{|c|}{Punjabi} & 23255 & 2836 & 5830 & 443 & 32364 \\ \hline
\multicolumn{1}{|c|}{Sanskrit} & 32385 & 1246 & 4006 & 265 & 37907 \\ \hline
\multicolumn{1}{|c|}{Tamil} & 16312 & 2803 & 5827 & 477 & 25419 \\ \hline
\multicolumn{1}{|c|}{Telugu} & 12078 & 2795 & 5776 & 442 & 21091 \\ \hline
\multicolumn{1}{|c|}{Urdu} & 22990 & 2801 & 5786 & 443 & 34280 \\ \hline
\end{tabular}%
}
\caption{Number of synsets in different wordnets}
\label{tab:iwnstats}
\end{table*}

\begin{table*}[h!]
\centering
\resizebox{0.75\textwidth}{!}{%
\begin{tabular}{c|r|r|r|r|r|r|r|r|r|}
\cline{2-10}
                                & \multicolumn{2}{c|}{Nouns}                      & \multicolumn{2}{c|}{Verbs}                      & \multicolumn{2}{c|}{Adjectives}                 & \multicolumn{2}{c|}{Adverbs}                    & \multicolumn{1}{c|}{\multirow{2}{*}{Total}} \\ \cline{2-9}
                                & \multicolumn{1}{c|}{D} & \multicolumn{1}{c|}{H} & \multicolumn{1}{c|}{D} & \multicolumn{1}{c|}{H} & \multicolumn{1}{c|}{D} & \multicolumn{1}{c|}{H} & \multicolumn{1}{c|}{D} & \multicolumn{1}{c|}{H} & \multicolumn{1}{c|}{}                       \\ \hline
\multicolumn{1}{|c|}{Assamese}  & 7019                   & 679                    & 1300                   & 36                     & 2744                   & 0                      & 294                    & 0                      & 12072                                       \\ \hline
\multicolumn{1}{|c|}{Bengali}   & 11049                  & 7680                   & 1824                   & 99                     & 3356                   & 3                      & 312                    & 0                      & 24323                                       \\ \hline
\multicolumn{1}{|c|}{Bodo}      & 6940                   & 603                    & 1594                   & 64                     & 2854                   & 1                      & 293                    & 0                      & 12349                                       \\ \hline
\multicolumn{1}{|c|}{Gujarati}  & 10910                  & 7533                   & 1825                   & 99                     & 3356                   & 3                      & 312                    & 0                      & 24038                                       \\ \hline
\multicolumn{1}{|c|}{Hindi}     & 11584                  & 8221                   & 1988                   & 212                    & 3542                   & 4                      & 344                    & 0                      & 25895                                       \\ \hline
\multicolumn{1}{|c|}{Kannada}   & 7806                   & 1973                   & 1921                   & 154                    & 3453                   & 3                      & 133                    & 0                      & 15443                                       \\ \hline
\multicolumn{1}{|c|}{Kashmiri}  & 9363                   & 6261                   & 1767                   & 100                    & 3240                   & 2                      & 294                    & 0                      & 21027                                       \\ \hline
\multicolumn{1}{|c|}{Konkani}   & 10545                  & 6952                   & 1888                   & 128                    & 3391                   & 2                      & 328                    & 0                      & 23234                                       \\ \hline
\multicolumn{1}{|c|}{Malayalam} & 9146                   & 4754                   & 1970                   & 206                    & 3525                   & 4                      & 340                    & 0                      & 19945                                       \\ \hline
\multicolumn{1}{|c|}{Manipuri}  & 7192                   & 823                    & 1324                   & 43                     & 2712                   & 0                      & 244                    & 0                      & 12338                                       \\ \hline
\multicolumn{1}{|c|}{Marathi}   & 9874                   & 6556                   & 1839                   & 144                    & 3092                   & 0                      & 333                    & 0                      & 21838                                       \\ \hline
\multicolumn{1}{|c|}{Nepali}    & 5217                   & 496                    & 1114                   & 42                     & 2202                   & 1                      & 200                    & 0                      & 9272                                        \\ \hline
\multicolumn{1}{|c|}{Odiya}     & 11039                  & 7680                   & 1679                   & 66                     & 3187                   & 2                      & 271                    & 0                      & 23924                                       \\ \hline
\multicolumn{1}{|c|}{Punjabi}   & 10215                  & 6382                   & 1822                   & 99                     & 3355                   & 3                      & 312                    & 0                      & 22188                                       \\ \hline
\multicolumn{1}{|c|}{Sanskrit}  & 8396                   & 6470                   & 1048                   & 28                     & 2873                   & 2                      & 241                    & 0                      & 19058                                       \\ \hline
\multicolumn{1}{|c|}{Tamil}     & 8130                   & 3066                   & 1821                   & 98                     & 3353                   & 3                      & 312                    & 0                      & 16783                                       \\ \hline
\multicolumn{1}{|c|}{Telugu}    & 6944                   & 1843                   & 1819                   & 98                     & 3350                   & 0                      & 312                    & 0                      & 14366                                       \\ \hline
\multicolumn{1}{|c|}{Urdu}      & 10424                  & 6816                   & 1822                   & 98                     & 3356                   & 3                      & 313                    & 0                      & 22832                                       \\ \hline
\end{tabular}%
}
\caption{Linkage Statistics for English to Indian Language WordNets. D stands for Direct links, and H stands for Hypernymy links}
\label{tab:linkstats}
\end{table*}

\section{Resources}
\label{resources}

In this section, we describe the resources released with our work. We release two primary resources with our dataset which are described in subsections \ref{subsec:ilw} and \ref{subsec:linkres} below. 
\subsection{Indian Language WordNets}
\label{subsec:ilw}

The creation of IndoWordNet began in 2000 with Hindi WordNet. Due to the complex nature of Indian language families, and many other reasons such as morphological richness, gender information etc. it was decided that Hindi be used as a pivot for linking all the Indian Languages. Hindi shares many common features and borrowed concepts from ancient Indian languages like Sanskrit and is the most commonly spoken language in India. The expansion approach adopted for IndoWordNet creation is:
\begin{enumerate}
\item Creation of a Hindi synset with synonymous words.
\item Mapping of the synset with relations such as hypernymy and hyponymy \textit{etc.}
\item Tagging of the synset with an ontological category.
\item Allotment of a unique synset ID to the concept described in the synset.
\item Creation of the same synset in the other Indian languages leading to an implicit linkage of relations, ontological categories.
\end{enumerate}

We release the latest data in IndoWordNet with statistics described in subsection \ref{subsec:currstats} below.

\subsubsection{Construction Principles}

\begin{itemize}
\item \textbf{Minimality:} We try to capture the minimal set of words in the synset which uniquely define the concept and ensure that it is identifiable via the use of these words.
\item \textbf{Coverage:} We also try to stress on the completion of the synset and try to capture all the words which represent the concept.
\item \textbf{Replaceability:} This principle states that all the words in the synset should be able to replace one another in an example sentence quoted along with the synset. These words must be able to replace each other in the same sense.
\end{itemize}

\subsubsection{Current Statistics: IndoWordnet}
\label{subsec:currstats}
Table \ref{tab:iwnstats} shows the statistics of the released wordnets. These wordnets have, on an average, approximately 28,000 synsets, with Nepali and Hindi having the minimum and the maximum number of synsets respectively. The number of synsets in Hindi is maximum due to the fact that work on IndoWordNet started with the Hindi language. It should also be noted that the ratio of nouns, verbs, adjectives, and adverbs is also on an average 48:6:13:1; the trend being similar to Princeton WordNet. 

\subsection{Linkage between English and Indian Language WordNets}
\label{subsec:linkres}

For linking Indian language wordnets with the Princeton WordNet, we link the Hindi Wordnet data with Princeton WordNet data manually with the help of lexicographers. This has been an ongoing work since many years, and a resource release was long standing. We delve deep into the language related issues in linking both the languages and ensure that only a valid relation is established between both the lexicons. The principles used and the current linkage statistics are described in the subsections below.
\subsubsection{Principles}
We use the simple principles of concept representation to ensure a valid linkage between the two languages. While linking two concepts, we refer to all words present in both the synsets for creating the linkage. First, we start with linking the known common concepts between both the WordNets of Hindi and English (Direct Linkages). We, then, start to link Hypernymy linkages from Hindi to English. For \textit{e.g.,} \textit{younger paternal uncle} and \textit{elder paternal uncle} are two different specific concepts, and thus have two different synsets in Hindi language. English language, on the other hand, has only the concept of \textit{uncle}, and hence we link both the Hindi language concepts to uncle as Hypernymy linkages.
\begin{figure}[ht]
\centering
\includegraphics[width=\linewidth]{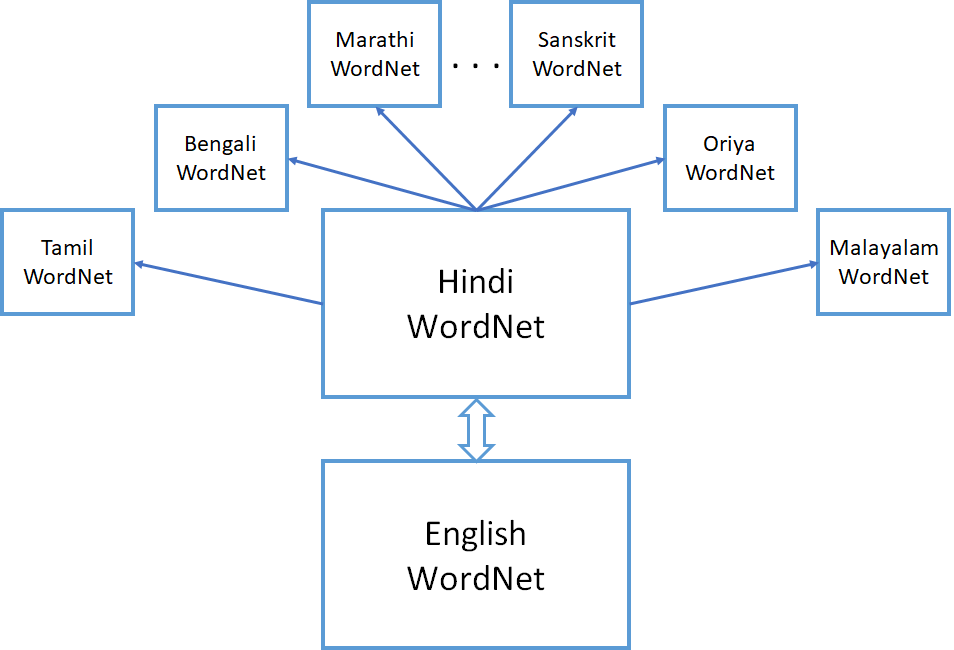}
\caption{Indian Language WordNet linkages with Princeton WordNet. D stands for links of the type Direct, whereas H stands for the links of the type HYPERNYM.}
\label{fig:wordnetLinks}
\end{figure}
\subsubsection{Princeton Statistics}

At present, the Princeton Wordnet has a total of 117659 synsets, with 82115 nouns, 13767 verbs, 18156 adjectives, and 3621 adverbs\footnote{\url{https://wordnet.princeton.edu/wordnet/man/wnstats.7WN.html}}. They further categorize some of their adjectives into satellite adjectives but the statistics shown include both adjectives and satellite adjectives. We use Princeton WordNet version 3.0 for the purpose of linkage. We began linking Hindi WordNet with version 2.1 and shifted to WordNet version 3.0 using the mappings provided\footnote{\url{https://wordnet.princeton.edu/man/sensemap.5WN.html}} by Princeton WordNet.

\subsubsection{Current Statistics: Linkages for Language pairs}
Table \ref{tab:linkstats} shows the statistics of the released linkages. There are approximately 20,000 links for an English-Indian language pair on average, with Nepali and Hindi having the minimum and the maximum number of links. Again, the number of links in Hindi is maximum due to the fact that work on IndoWordnet started with the Hindi language, and we link Hindi directly with English. At times, the concept present in Hindi is not present in the other Indian languages thus leading to the less number of linkages for the other languages, in some cases. Table \ref{tab:linkstats} show part-of-speech category-wise distribution of the linked synsets, and also indicated the number of directly linked synsets (D) along with the synset linkages which have been marked as hypernymy linkages (H).

The statistics show our progress in updating IndoWordnet as a resource. The relatively large number of linkages also show that the Indian wordnets have matured considerably.
\section{Discussion}
\label{disc}
Many concepts in the Indian languages are specific to the Indian culture. Thus, their corresponding variant is not available in the Princeton WordNet (and is not likely to be included anytime). Thus, one needs to maintain the translation/transliteration of such notions from Indian languages to the English language as a separate bilingual mapping \footnote{Since bilingual mappings are not standardized, we do not release them along with our resources}.  A similar issue arises in case of proper nouns, which should be present in an Indian lexicon but they are not present in Princeton WordNet. They are also handled using bilingual mappings \cite{singh2016mapping}. Some of the synsets in Indian languages are too fine-grained and have a common representation in the English language. This is why we use the principle of Hypernymy linkages for linking such concepts. We reserve a set of synset id numbers later for language specific concepts and create them to include in these wordnets, individually. These are not linked to the Princeton WordNet and hence are not included in our resource.

\section{Conclusion and Future Work}

In this paper, we describe two resources released along with this paper. We discussed the Indian language wordnets that are part of the IndoWordNet project. We enlisted the statistics of the latest version, which we provide as a single bundle along with this paper. Next, we described the linkage process for creating English-Indian language links using English-Hindi language links. We then enlisted the statistics of the latest version of this linked data, which is also provided along with this paper.

In future, we plan to continue building the wordnets and increase linkage. We will also investigate semi-automatic linkage tools such as the ones created by \newcite{2012PB}, \textit{etc.} so that the workload on our lexicographers and researchers can be reduced to a certain extent


\section{Acknowledgement}

We would like to acknowledge the work done by lexicographers at Center For Indian Language Technology (CFILT), IIT Bombay without which we would not have been able to link Indian Language Wordnets.

\nocite{*}
\section{Bibliographical References}
\label{main:ref}

\bibliographystyle{lrec}
\bibliography{xample}


\end{document}